%% file: main.tex
\documentclass[10pt,twocolumn,letterpaper]{article}
\input{env.tex}

\begin{document}

\title{SATR: Zero-Shot Semantic Segmentation of 3D Shapes}

\author{Ahmed Abdelreheem$^{1}$\quad Ivan Skorokhodov$^{1}$\quad Maks Ovsjanikov$^{2}$\quad Peter Wonka$^{1}$\quad\\
$^{1}$KAUST \quad $^{2}$LIX, Ecole Polytechnique\\
\tt \small \{asamirh.95,iskorokhodov,pwonka\}@gmail.com, \tt \small \{maks\}@lix.polytechnique.fr
}

\input{figures/teaser}

\input{sections/0_Abstract.tex}



\input{sections/1_Introduction.tex}
\input{sections/2_RelatedWork.tex}
\input{sections/3_Method.tex}

\input{sections/4_Experiments.tex}
\input{sections/5_Conclusion.tex}


{\small
\bibliographystyle{ieee_fullname}
\bibliography{egbib}
}

\clearpage
\appendix
\renewcommand\thefigure{\arabic{figure}}    
\setcounter{figure}{0}
\setcounter{table}{0}
\input{appendices/limitations}

\input{appendices/implementation-details}

\input{appendices/qualitative-results}
\input{appendices/ablation-studies}

\end{document}

%% file: env.tex
\usepackage{iccv}
\usepackage[accsupp]{axessibility}  
\usepackage{times}
\usepackage[pdftex]{graphicx}
\usepackage{epsfig}
\usepackage{amsmath}
\usepackage{amssymb}
\usepackage{url}
\usepackage{amsmath} 
\usepackage{amssymb} 
\usepackage{multirow}
\usepackage{makecell} 
\usepackage{subcaption} 
\usepackage{bm} 
\usepackage{stmaryrd} 
\usepackage{dsfont} 
\usepackage{pifont} 
\usepackage{ifthen} 
\usepackage{booktabs} 
\usepackage{soul}
\usepackage{wrapfig} 
\usepackage{colortbl} 

\usepackage[ruled,vlined]{algorithm2e} 

\usepackage[pagebackref=true,breaklinks=true,bookmarks=false,colorlinks]{hyperref}
\usepackage[capitalize]{cleveref}

\iccvfinalcopy 

\def\iccvPaperID{****} 

\ificcvfinal\fi

\definecolor{mediumtealblue}{rgb}{0.0, 0.33, 0.71}
\definecolor{darkpastelgreen}{rgb}{0.01, 0.75, 0.24}
\definecolor{azure}{rgb}{0.0, 0.5, 1.0}

\newcommand{\apref}[1]{\ref*{#1}}
\newcommand{\detector}{\mathsf{D}}

\newcommand{\R}{\mathbb{R}}
\newcommand{\mesh}{\mathcal{F}}
\newcommand{\prompt}{\bm t}
\newcommand{\image}{\bm x}
\newcommand{\score}{p}
\newcommand{\area}{s}
\newcommand{\weights}{\mathcal{W}}
\newcommand{\bbox}{\bm b}
\newcommand{\face}{\bm f}
\newcommand{\distset}{\bm d}
\newcommand{\vis}{\bm v}
\newcommand{\neighbours}{\mathcal{N}}
\newcommand{\vertbar}{\ |\ } 

\newcommand{\highlight}[1]{\textcolor{azure}{#1}}
\newcommand{\best}[1]{\textbf{#1}}

\newcommand{\projecturl}{https://samir55.github.io/SATR/}

\newcommand{\methodfullname}{Segmentation Assignment with Topological Reweighting}
\newcommand{\methodname}{SATR}

\newcommand{\expect}[2][]{
\ifthenelse{\equal{#1}{}}{
\mathbb{E}\left[#2\right]
}{
\underset{#1}{\mathbb{E}}\left[#2\right]
}}

\crefname{section}{Sec.}{Secs.}
\Crefname{section}{Section}{Sections}
\Crefname{table}{Table}{Tables}
\crefname{table}{Tab.}{Tabs.}

\newcommand{\tabincell}[2]{\begin{tabular}{@{}#1@{}}#2\end{tabular}}

%% file: figures/teaser.tex
\twocolumn[{
\maketitle
\ificcvfinal\thispagestyle{empty}\fi
\begin{center}
\captionsetup{type=figure}
\centering
\includegraphics[scale=1,width=\linewidth]{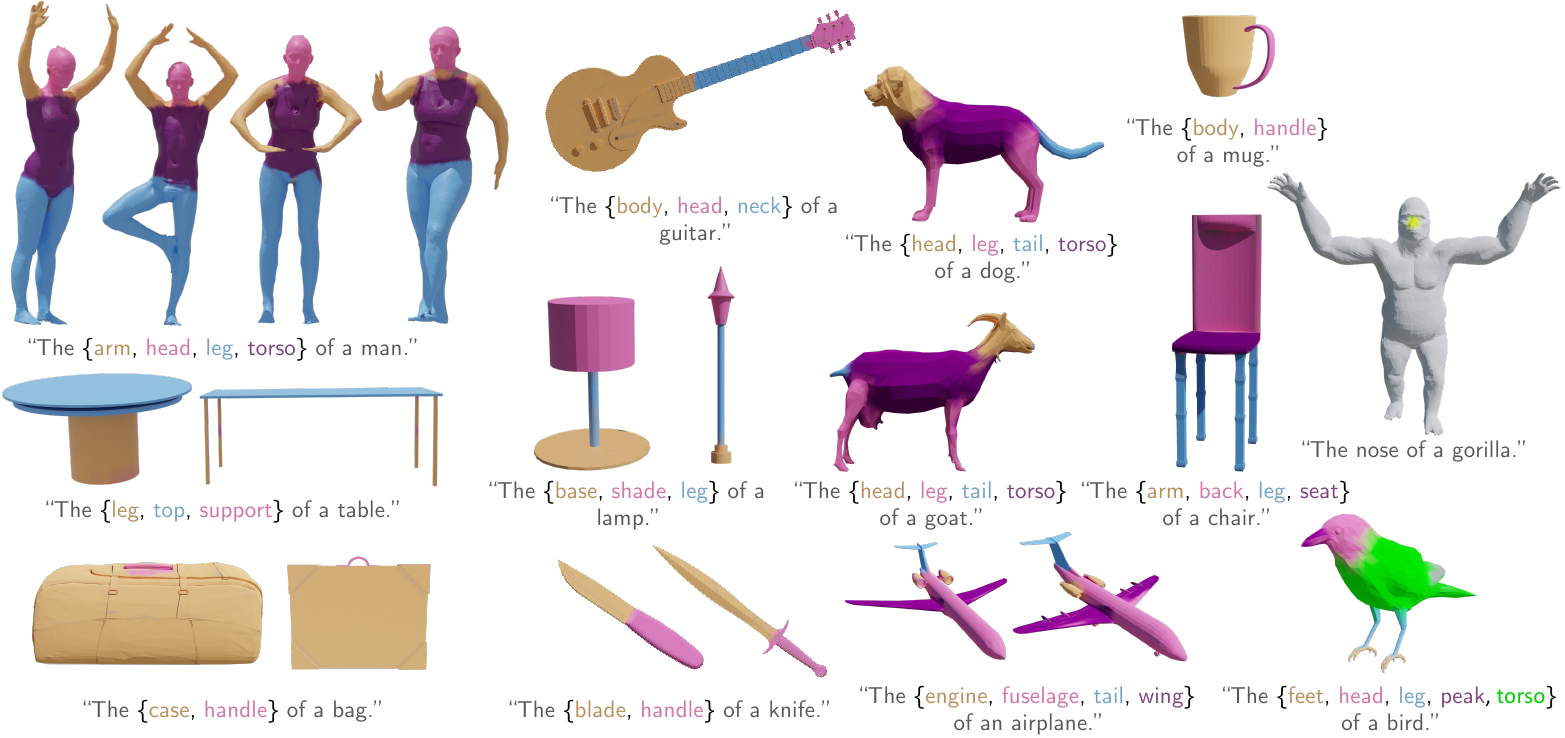}
\captionof{figure}{\methodname\ performs zero-shot 3D shape segmentation via text descriptions by using a zero-shot 2D object detector. It infers 3D segmentation from multi-view 2D bounding box predictions by exploiting the topological properties of the underlying surface. \methodname\ is able to segment the mesh from both single and multiple queries and provides accurate predictions even for fine-grained categories.}
\label{fig:teaser}
\end{center}
}]

%% file: sections/0_Abstract.tex
\begin{abstract}

We explore the task of zero-shot semantic segmentation of 3D shapes by using large-scale off-the-shelf 2D image recognition models.
Surprisingly, we find that modern zero-shot 2D object detectors are better suited for this task than contemporary text/image similarity predictors or even zero-shot 2D segmentation networks.
Our key finding is that it is possible to extract accurate 3D segmentation maps from multi-view bounding box predictions by using the topological properties of the underlying surface.
For this, we develop the \methodfullname\ (\methodname) algorithm and evaluate it on ShapeNetPart and our proposed FAUST benchmarks.
\methodname\ achieves state-of-the-art performance and outperforms a baseline algorithm by 1.3\% and 4\% average mIoU on the FAUST coarse and fine-grained benchmarks, respectively, and by 5.2\% average mIoU on the ShapeNetPart benchmark.
Our source code and data will be publicly released.
Project webpage: \projecturl.
\end{abstract}

%% file: sections/1_Introduction.tex
\section{Introduction}\label{sec:introduction}

Recent developments in vision-language learning gave rise to many 2D image recognition models with extreme zero-shot generalization capabilities (e.g., \cite{CLIP, SLIP, GLIP, LSeg, X-Decoder}).
The key driving force of their high zero-shot performance was their scale~\cite{GPT3}: both in terms of the sheer amount of data~\cite{LAION, OpenImagesV5} and parameters~\cite{Flamingo, ScalingVisionTransformers} and in terms of developing the architectures with better scalability~\cite{Transformer, ViT, SwinT}.
However, extending this success to the 3D domain is hindered by the limited amount of available 3D data~\cite{abdelreheem2022scanents,referit3d}, and also the higher computational cost of the corresponding architectural components~\cite{3D_Unet}.
For example, the largest openly available 2D segmentation dataset~\cite{OpenImagesV7} contains \textit{two} orders of magnitude more instance annotations than the largest 3D segmentation one~\cite{ScanNet200}.
This forces us to explore other ways of performing zero-shot recognition in 3D, and in our work, we explore the usage of off-the-shelf 2D models for zero-shot 3D shape segmentation.

Zero-shot 3D shape segmentation is a recently emerged research area~\cite{3D_highlighter} with applications in text-based editing~\cite{DreamFusion,shapetalk}, stylization~\cite{text2mesh}, and interactive visualization.
Given a 3D mesh, the user provides one or several text descriptions of their regions of interest, and the task is to categorize each face on the mesh into one of the given descriptions (or ``background'' class if it does not suit any).
To the best of our knowledge, the only previous work which explores this task is 3D Highlighter (3DH)~\cite{3D_highlighter}.
The method uses an optimization-based search algorithm guided by CLIP~\cite{CLIP} to select the necessary faces for a given text prompt.
While showing strong zero-shot performance, 3DH has two drawbacks: 1) it struggles in fine-grained segmentation, and 2) it is very sensitive to initialization (see \Cref{fig:3dhRandomSeeds}).
Moreover, due to its per-query optimization, the segmentation process is slow, taking up to ${\approx}$5-10 minutes on a recent GPU for a single semantic part.

In our work, we explore modern zero-shot 2D object detectors~\cite{GLIP} and segmentors~\cite{LSeg, CLIPSeg} for 3D shape segmentation.
Intuitively, 2D segmentation networks are a natural choice for this task: one can predict the segmentations for different views,  and then directly propagate the predicted pixel classes onto the corresponding mesh faces.
Moreover and surprisingly, we found that it is possible to achieve substantially higher performance using a zero-shot 2D object detector~\cite{GLIP}.
To do this, we develop \methodfullname\ (\methodname): a method that estimates a 3D segmentation map from multi-view 2D bounding box predictions by using the topological properties of the underlying 3D surface.
\input{figures/clip_initializations}

For a given mesh and a text prompt, our method first uses GLIP~\cite{GLIP} to estimate the bounding boxes from different camera views.
However, relying exclusively on the bounding boxes provides only coarse guidance for 3D segmentation and is prone to ``leaking'' unrelated mesh faces into the target segment.
This motivates us to develop two techniques to infer and refine the proper segmentation.
The first one, \textit{gaussian geodesic reweighting}, performs robust reweighting of the faces based on their geodesic distances to the potential segment center.
The second one, \textit{visibility smoothing}, uses a graph kernel, which adjusts the inferred weights based on the visibility of its neighbors.
When combined together, these  techniques allow for achieving state-of-the-art results on zero-shot 3D shape segmentation, especially for fine-grained queries.

To the best of our knowledge, there are currently no quantitative benchmarks proposed for 3D mesh segmentation, and all the evaluations are only qualitative~\cite{3D_highlighter}.
For a more robust evaluation, we propose one quantitative benchmark, which includes coarse and fine-grained mesh segmentation categories. We also evaluate our method on ShapeNetPart \cite{Yi16} benchmark.
Our proposed benchmark is based on FAUST~\cite{FAUST}: a human body dataset consisting of 100 real human scans.
We manually segment 17 regions on one of the scans and use the shape correspondences provided by FAUST to propagate them to all the other meshes.
We evaluate our approach along with  existing methods on the proposed benchmarks and show the state-of-the-art performance of our developed ideas both quantitatively and qualitatively.
Specifically, \methodname\ achieves 82.46\% and 46.01\% average mIoU on the coarse and fine-grained FAUST benchmarks and 31.9\% average mIoU scores on the ShapeNetPart benchmark, outperforming recent methods.
For fine-grained categories, the advantage of our method is even higher: it surpasses a baseline method by at least 4\% higher mIoU on average.
We will publicly release our source code and benchmarks.

%% file: figures/clip_initializations.tex
\begin{figure}
    \centering
    \includegraphics[width=\linewidth]{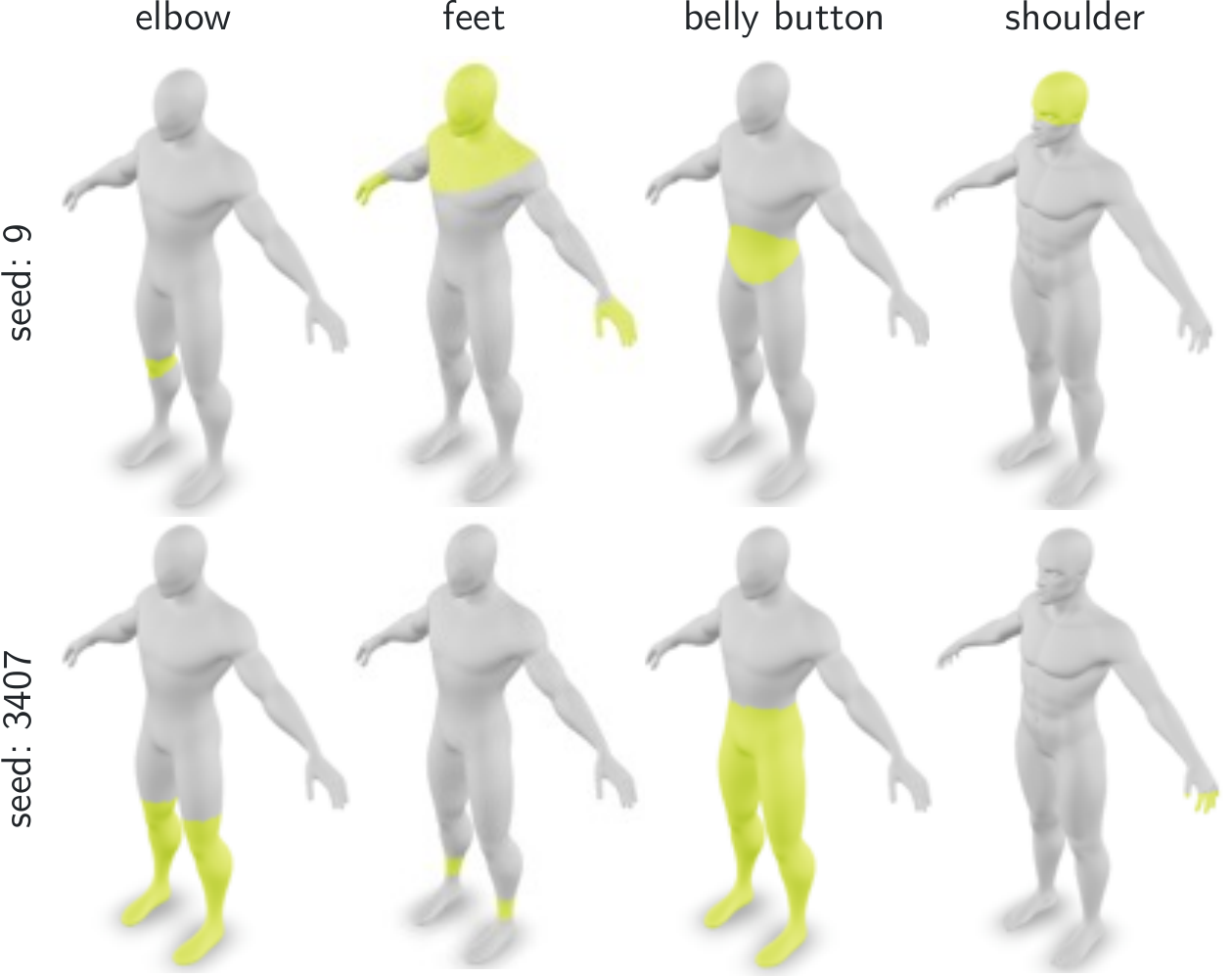}
    \caption{{3DHighlighter \cite{3dhighlighter} is very sensitive to initialization.} We observe that 3DHighlighter produces quite different results when using different seeds for the same prompt on different 3D shapes.}
    \label{fig:3dhRandomSeeds}
\end{figure}

%% file: sections/2_RelatedWork.tex
\section{Related Work}\label{sec:related-work}

\textbf{Zero-shot 2D detection and segmentation}.
Zero-shot 2D object detection is a fairly established research area~\cite{ZSD, ZSOD}.
Early works relied on pre-trained word embeddings~\cite{word2vec, Glove} to generalize to unseen categories (e.g., \cite{ZSD, ZSOD, ZSOD_hybrid_reg_emb, ContrastZSD}).
With the development of powerful text encoders~\cite{BERT} and vision-language multi-modal networks~\cite{CLIP}, the focus shifted towards coupling their representation spaces with the representation spaces of object detectors (e.g., ~\cite{ViLD, ProposalCLIP}).
The latest methods combine joint text, caption, and/or self-supervision to achieve extreme generalization capabilities to unseen categories (e.g, ~\cite{GLIP, GLIP2, GRiT, X-Decoder, OVDet, ObjectCentricOVD, ClassAgnosticOD,gadre2022cows,dong2023maskclip}).

\begin{figure}
    \centering
    \includegraphics[width=\linewidth]{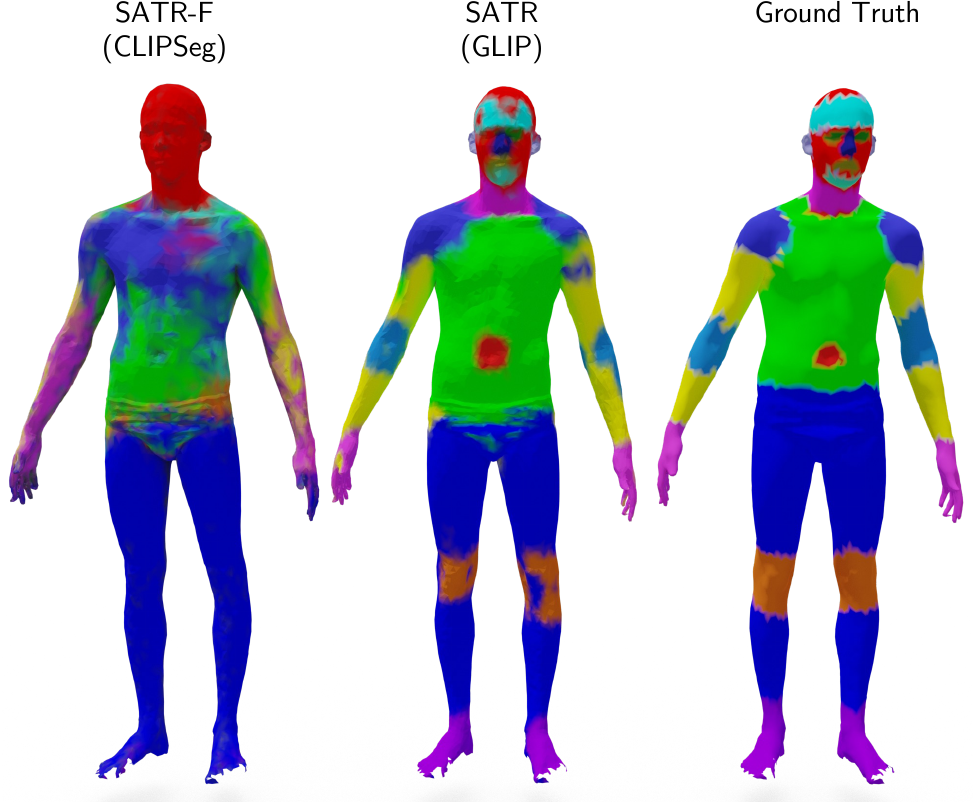}
    \caption{CLIPSeg \cite{CLIPSeg} struggles to identify fine-grained parts compared to GLIP, which is a detection-based method. We show the segmented shapes using each of CLIPSeg and GLIP as a backbone for our proposed algorithm SATR. The textual prompts consist of all 17 semantic regions.}
    \label{fig:clipseg}
\end{figure}

Zero-shot 2D segmentation is a more challenging problem~\cite{X-Decoder} but had a similar evolutionary path to zero-shot detection.
Early works (e.g., \cite{ZS3Net, CaGNet, JoEm, OVSP, SIGN, SPNet, STRICT, PMOSR}) focused on adapting open-vocabulary word embeddings to align the semantic representations with the image recognition model.
Later ones (e.g., \cite{ZegFormer, Fusioner, LSeg, CLIPSeg, MaskCLIP, OpenSeg, SegCLIP, SimpleZ3S, SAN-CLIPSeg}) developed the mechanisms to adapt rich semantic spaces of large-scale multi-modal neural networks (e.g., \cite{CLIP, SLIP, ALIGN}) for segmentation tasks.
Some recent works show that one can achieve competitive segmentation performance using only text or self-supervision~\cite{GroupViT, DINO, Peekaboo, MV-SimCon, TCL}.
The current state-of-the-art zero-shot 2D semantic segmentation models are based on rich multi-modal supervision (e.g., \cite{OVSegmentor, GLIP2, X-Decoder}).
A limiting factor of 2D segmentation is the absence of large-scale segmentation datasets due to the high annotation cost~\cite{RefCOCO, MS-COCO, OpenImagesV4, OpenImagesV5}, which hinders generalization.
However, we observed that these models struggle to produce fine-grained segmentation even in 2D, especially for fine-grained categories (see \Cref{fig:clipseg}) and develop an algorithm that constructs accurate segmentation predictions from estimated bounding boxes.

\textbf{Zero-shot 3D segmentation}.
Zero-shot 3D segmentation is a new research topic, and the main focus of the community was targeted towards point cloud segmentation~\cite{3DGenZ, ZSPCSeg_GeomPrim, PLA, PartSLIP,PartGlot}.
With the rise of Neural Radiance Fields (NeRFs)~\cite{NeRF, NeuralVolumes}, there were several methods developed to model \textit{semantic} fields(e.g., \cite{SemanticNeRF, NeSF, NeRF-SOS, PanopticNeRF, PanopticNeRFUrban, NFFF, PanopticLifting,hong2023threedclr,abdelreheem2023zeroshot}) by reconstructing ground-truth or estimated semantic annotations from multi-view renderings.
By distilling zero-shot 2D segmentation networks (e.g., \cite{LSeg, DINO}) into a NeRF, these methods can perform 3D segmentation of a volumetric scene (e.g., \cite{DFF, CLIP-Fields, ISRF}) and can generalize to an open-set vocabulary.
By fusing representations from additional feature extractors of non-textual modalities, ConceptFusion~\cite{ConceptFusion} can also support zero-shot visual and audio queries.
In our case, we are interested in shape segmentation and show that employing a 2D object detector yields state-of-the-art results.

PartSLIP~\cite{PartSLIP} is concurrent work that performs zero/few-shot part segmentation of point clouds and, similarly to us, also relies on the GLIP~\cite{GLIP} model.
It clusters a point cloud, predicts bounding boxes via GLIP for multiple views, and assigns a score to each cluster depending on the number of its visible points inside each bounding box. Their method is designed for point clouds while ours is designed for meshes.

To the best of our knowledge, the only prior work which explores zero-shot 3D mesh segmentation is 3D Highlighter (3DH)~\cite{3D_highlighter}.
It solves the task by optimizing a probability field of a point to match a given text prompt encoded with CLIP~\cite{CLIP}.
While showing strong generalization capabilities, their approach struggles to provide fine-grained predictions and is very sensitive to initialization (see \Cref{fig:3dhRandomSeeds}).

\input{figures/method}

%% file: figures/method.tex
\begin{figure*}[!htb]
    \centering
    \includegraphics[width=0.86\linewidth]{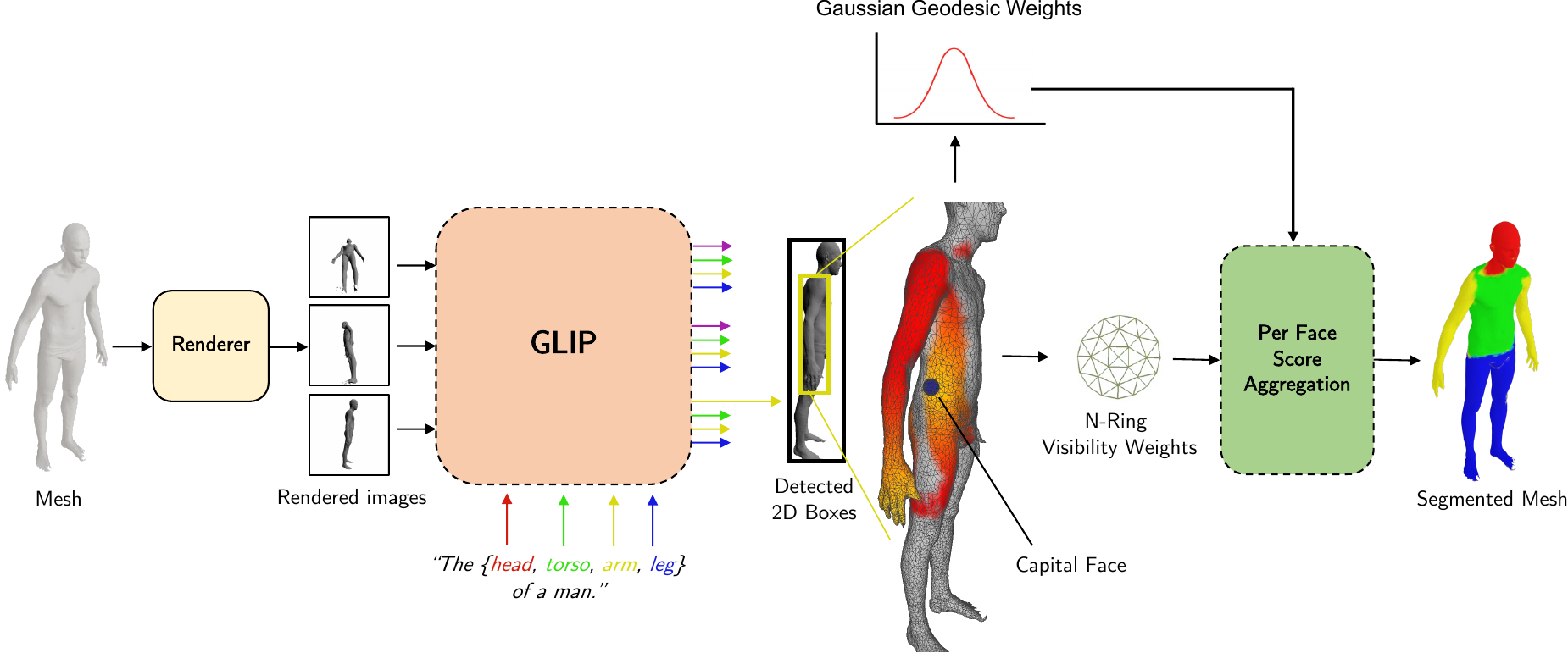}
    \caption{An overview of our method. Meshes are rendered from random viewpoints. The resulting images are processed by GLIP, which detects bounding boxes in the images. Each bounding box corresponds to a prompt (segment). For each bounding box, we compute scores for triangles inside the bounding box using \emph{Gaussian Geodesic Reweighting} and \emph{Visibility Smoothing}. Aggregating the scores yields a segmented mesh.  }
    \label{fig:method}
\end{figure*}

%% file: sections/3_Method.tex
\section{Method}\label{sec:method}

As input, we assume a 3D shape, represented as a polygon mesh $\mesh = \{ \face_n \}_{i=n}^N$ of $d$-sided polygon faces $\face_n \in \R^{d \times 3}$, and $K$ semantic text descriptions $\{\prompt_k\}_{k=1}^K$, provided by the user. 
The prompts $\prompt_k$ are single nouns or compound noun phrases consisting of multiple words.
Then, the task of 3D shape segmentation is to extract $K$ non-intersecting subsets $\{\mesh_k \vertbar \mesh_k \subseteq \mesh \}_{k=1}^K$ in such a way that the $k$-th subset $\mesh_k$ is a part of the mesh surface which semantically corresponds to the $k$-th text prompt $\prompt_k$.

In our work, we explore modern powerful 2D vision-language models to solve this task.
The simplest way to incorporate them would be employing a zero-shot 2D segmentation network, like CLIPSeg~\cite{CLIPSeg} or LSeg~\cite{LSeg}, which can directly color the mesh given its rendered views.
But as we show in experiments, this leads to suboptimal results since modern zero-shot 2D segmentors struggle to produce fine-grained annotations and high-quality segments, while 2D detectors could be adapted for shape segmentation with surprisingly high precision.

In our work, we consider untextured meshes, but it is straightforward to apply the method to textured ones.

In this way, our method relies on a zero-shot 2D object detector $\detector(\image, \prompt) = \{(\bm b_\ell, \score_\ell)\}_{\ell=1}^L$, which takes as input an RGB image $\image \in \R^{H \times W \times 3}$ of size $H \times W$ and a text prompt $\prompt$, and outputs $L \geq 0$ bounding boxes $\bbox_\ell = (x_\ell, y_\ell, h_\ell, w_\ell) \in \R^4$ with their respective probability scores $\score_\ell \in [0, 1]$.
In our notation, each bounding box $\bbox_\ell$ is defined by its left lower corner coordinate $(x_\ell, y_\ell)$, width $w_\ell$, and height $h_\ell$.
We choose GLIP~\cite{GLIP} as the detector model due to its state-of-the-art generalization capabilities.

In this section, we first describe a topology-agnostic baseline framework (denoted as SATR-F) that can leverage a 2D object detector to segment a meshed 3D surface.
Then, we explain why it is insufficient to infer accurate segmentation predictions (which is also confirmed by the experiments in \Cref{tab:faustfg,tab:faustcg,tab:shapenetpart_results}) due to the coarse selection nature of the bounding boxes.
After that, we develop our \methodfullname\ (denoted as SATR (F+R) or \methodname{} in short) algorithm, which alleviates this shortcoming by using the topological properties of the underlying surface. 

\subsection{Topology-Agnostic Mesh Segmentation}\label{sec:method:baseline}

Our topology-agnostic baseline method works the following way.
We render the mesh $\mesh$ from $M$ random views (we use $M = 10$ in all the experiments, if not stated otherwise) to obtain RGB images $\image_m$. To create the $M$ views, we generate random camera positions where the elevation and azimuth angles are sampled using the normal distribution ($\mu=0.7$,$\sigma=4$). We use this view generation to be directly comparable to 3DHighlighter \cite{3D_highlighter}. After that, for each view $\image_m$ and for each text prompt $\prompt_k$, we use the detector model to predict the bounding boxes with their respective confidence scores:
\begin{equation}\label{eq:detector-preds}
\{(\bbox_\ell^{m,k}, \score_{\ell}^{m,k})\}_{\ell=1}^{L_{m,k}} = \detector(\bbox_m, \prompt_k).
\end{equation}

Then, we use them to construct the initial face weights matrix $\weights'_m \in [0, 1]^{N \times K}$ for the $m$-th view.
Let $\mesh_\ell^{m,k} \subseteq \mesh$ denote a subset of visible (in the $m$-th view) faces with at least one vertex inside bounding box $\ell^{m,k}$, whose projection falls inside $\bbox_\ell^{m, k}$.
Then

\begin{equation}
\weights_m[n, k] = \sum_{\ell=1}^{L_{m, k}}{\weights^{\ell}_m[n, k, \ell]}
\end{equation}

\begin{equation}
\weights^{\ell}_m[n, k, l] = \begin{cases}
\score_{\ell}^{m,k}  &~\text{if $\face_n \in \mesh_\ell^{m,k}$} \\
0 &~\text{otherwise.}
\end{cases}
\end{equation}

In this way, the score of each face $\face_n$ for the $m$-th view is simply set to the confidence $\score_{\ell}^{m,k}$ of the corresponding bounding box(es) it fell into.

The face visibility is determined via the classical $Z$-culling algorithm \cite{KaolinLibrary}. 
In this way, if there are no bounding boxes predicted for $\bm x_m$ for prompt $\prompt_k$, then $\mathcal{W}_m[n, k]$ equals the zero matrix.
The latter happens when the region of interest is not visible from a given view or when the detector $\detector$ makes a prediction error.

Next, we take into account the area sizes of each face projection.
If a face occupies a small area inside the bounding box, then it contributed little to the bounding box prediction.
The area $\area_{m,\ell}^n$ of the face $\face_n$ in the view $\image_m$ and bounding box $\ell$ is computed as the number of pixels which it occupies.
We use the computed areas to re-weight the initial weights matrix $\weights'_m$ and obtain the area-adjusted weights matrix $\weights_m \in \R^{N \times K}$ for the $m$-th view:
\begin{equation}
\weights_m[n, k] = \sum_{\ell=1}^{L_{m, k}}{\weights^{\ell}_m[n, k, \ell] \times \area_{m,\ell}^n}
\end{equation}

To compute the final weights matrix, we first aggregate the predictions from each view $\image_m$ by summing the scores of the un-normalized matrix $\tilde{\weights} \in \R^{N \times K}$:
\begin{equation}
\tilde{\weights}[n, k] = \sum_{m}{\weights_m[n, k]},
\end{equation}
and then normalize it by dividing each column by its maximum value to obtain our final weights matrix $\weights \in \R^{N \times K}$:
\begin{equation}
\weights[n, k] = \tilde{\weights}[n, k] / \max_k \tilde{\weights}[n, k].
\end{equation}

The above procedure constitutes our baseline method of adapting bounding box predictions for 3D shape segmentation.
As illustrated in \Cref{fig:leaking}, its disadvantage is ``\textit{segmentation leaking}'': some semantically unrelated parts of the surface get incorrectly attributed to a given text prompt $\prompt$, because they often fall into predicted bounding boxes from multiple views.
To alleviate this issue, we develop a more careful score assignment algorithm that uses the topological properties of the surface, thus allowing us to obtain accurate 3D shape segmentation predictions from a 2D detector.

\subsection{Gaussian Geodesic Reweighting}\label{sec:method:geodesic}

Bounding box estimates give only coarse estimates about the semantic region being queried, and we found that using the surface topology allows localizing it with substantially better precision.
For this, we construct a method that utilizes geodesic distances between mesh faces, i.e., path lengths from one face to another along the surface, instead of their direct distances in Euclidian space.

Consider the following example of segmenting a human palm.
When the hand is in a resting state, the palm lies close to the waistline in Euclidean space (as illustrated in \Cref{fig:normalizationheatmaps}).
Then, a simple topology-agnostic method would lead to the waistline leaking into the segmentation prediction.
At the same time, one can observe that the predicted bounding boxes are always centered around the palm, and the waistline is far away from it in terms of the geodesic distance.
In this way, discarding such outlying polygons yields a precise segmentation of the required region.
And this is the main intuition of our developed algorithm.

As a first step, for each predicted bounding box $\bbox_{\ell}^{m,k}$, we estimate its central face, which we call the \textit{capital face} $g^{m,k}_\ell \in \mesh$.
It is computed by taking the (area-weighted) average of all the vertices from all the faces inside $\mesh_{\ell}^{m,k}$, projecting this average point onto $\mesh$ and taking the face on which the projection lies.
After that, we compute a vector of geodesic distances $\distset_\ell^{m,k} \in \R_+^{N}$ from the capital face $g^{m,k}_\ell$ to every other face $f \in \mesh_\ell^{m,k}$:
\begin{equation}
\distset_\ell^{m,k}[n] = \begin{cases}
\text{gdist}(g_\ell^{m,k}, \face_n) &\text{if~} \face_n \in \mesh_\ell^{m,k} \\
0 &\text{otherwise}
\end{cases}
\end{equation}
where $\text{gdist}(\cdot, \cdot)$ denotes the geodesic length between two faces computed on the mesh $\mesh$ using the Heat method~\cite{Heat_method}.

It feels natural to use those geodesic distances directly to reweight the original weight matrix $\weights$.
However, this leads to sub-optimal results for two reasons: 1) there are natural errors in selecting the capital face, which would bias reweighting towards incorrect regions; and 2) as we show in \Cref{fig:normalizationheatmaps}, it is difficult to tune the decay rate for such reweighting.
Instead, we propose \textit{Gaussian reweighting} and demonstrate that it is more robust and accurate in practice. 
It works the following way.

First, we fit a Gaussian distribution over the distances and compute the corresponding probability density values for each face given its geodesic distance from the capital face:
\begin{equation}
\bm r_\ell^{m,k} \triangleq \{ \mathcal{N}(d; \mu_\ell^{m,k}, (\sigma_\ell^{m,k})^2) \vertbar d \in \distset_\ell^{m,k} \},
\end{equation}
where $\mu_\ell^{m,k}, \sigma_\ell^{m,k}$ denote the mean and standard deviation of the distances $\distset_\ell^{m,k}$.
This formulation nudges the weights away from the capital and works like adaptive regularization.
If there are inaccuracies in the capital face selection, then it will have less influence on the segmentation quality.
We aggregate the weights from multiple views into a single vector of scores $\bm r^{m,k} \in \R_+^N$ and reweigh the original weight matrix $\weights_m$ for the $m$-th view to obtain the weight matrix $\weights_m^g$ with Gaussian geodesic reweighting: 
\begin{equation}
\weights_m^g[n,k] = \sum_{\ell=1}^{L_{m, k}}{\weights^{\ell}_m[n, k, \ell] \times \area_{m,\ell}^n \times \bm r_\ell^{m,k}}.
\end{equation}

After that, we compute the final weight matrix $\weights^g \in \R^{N\times K}$ for each face $\face_n \in \mesh$ in a similar manner to $\weights$ by taking the summing over $\weights_m^g$ from different views. We find that not normalizing the weight matrix whenever the Gaussian Geodesic Reweighting is used yields better performance than applying normalization.

This procedure takes into account the topological structure of the mesh surface, which allows for obtaining more accurate segmentation performance, as can be seen from \Cref{tab:faustcg,tab:faustfg} and \Cref{fig:qualitative}.
However, it has one drawback: it pushes the weights away from the capital face, which might decrease the performance.
To compensate for this, we develop the technique of \textit{visibility smoothing}, which we describe next.

\input{figures/heatmap}

\subsection{Visibility Smoothing}\label{sec:method:smoothing}

The score matrix $\weights^g$ with Gaussian geodesic reweighting might allocate too little weight on the central faces around the capital face $g_\ell^{m,k}$, which happens for regions with a large average distance between the faces.
To compensate for this, we propose \textit{visibility smoothing}.
It works independently in the following way.

For each visible face $\face \in \mesh_\ell^{m,k}$, we compute its local neighborhood, where the neighbors are determined via mesh connectivity: a face $\bm g$ is a neighbor of face $\face$ if they share at least one vertex.
For this, we use a $q$-rank neighborhood $\neighbours_q(\face)$ (we use $q=5$ in all the experiments unless stated otherwise), which is constructed the following way.
For face $\face_n \in \mesh$, we take the face $\bm g \in \mesh$ if there exists a path on a graph between $\face$ and $\bm g$ of at most $q$ other vertices.

After that, we compute the neighborhood visibility scores vector $\vis_\ell^{m,k} \in [0,1]^N$ for each face $\bm f_n \in \mesh$ by computing the ratio between visible faces in its neighborhood and the overall neighborhood size:
\begin{equation}
\vis_\ell^{m,k}[n] = \frac{|\neighbours_q(\face_n) \cap \mesh_\ell^{m,k}|}{|\neighbours_q(\face_n)|}.
\end{equation}
Similarly to geodesic weighting, we aggregate the neighborhood visibility scores $\vis_\ell^{m,k}[n]$ across the bounding boxes into $\vis^{m,k}[n] \in [0,1]^N$ via element-wise vector summation:
\begin{equation}
\vis^{m,k}[n] = \sum_{\ell=1}^L \vis^{m,k}_\ell[n]
\end{equation}
This gives us our final per-view score matrix $\weights_m^* \in \R_+^{N \times K}$:
\begin{equation}
\weights_m^*[n, k] = \sum_{\ell=1}^{L_{m, k}}{\weights^{\ell}_m[n, k, \ell] \times \area_{m,\ell}^n \times \bm r_\ell^{m,k} \times \bm \vis^{m,k}[n]}. 
\label{eq:face_score_computation}
\end{equation}
Again, we aggregate our multi-view scores $\weights_m^*[n]$ into the final weights matrix $\weights^*[n]$ by taking the maximum across the views.

We call the above technique \textit{visibility smoothing} since it smoothes face weights according to their neighborhood visibility and can be seen as a simple convolutional kernel over the graph.
It allows for repairing the weights in the central part of the visible surface region without damaging the rest of the faces.
This consequently leads to a noticeable improvement in the scores, as we report in \Cref{tab:ablationsFaustCG,tab:faustfg}.

The pseudo-code of our algorithm is provided in Algorithm~\ref{alg:satr} in Appx~\apref{ap:pseudocode}, together with additional implementation details.
Also, note that the source code of our method will be publicly released.

%% file: figures/heatmap.tex
\begin{figure}[!htb]
    \centering
    \includegraphics[width=0.9\linewidth]{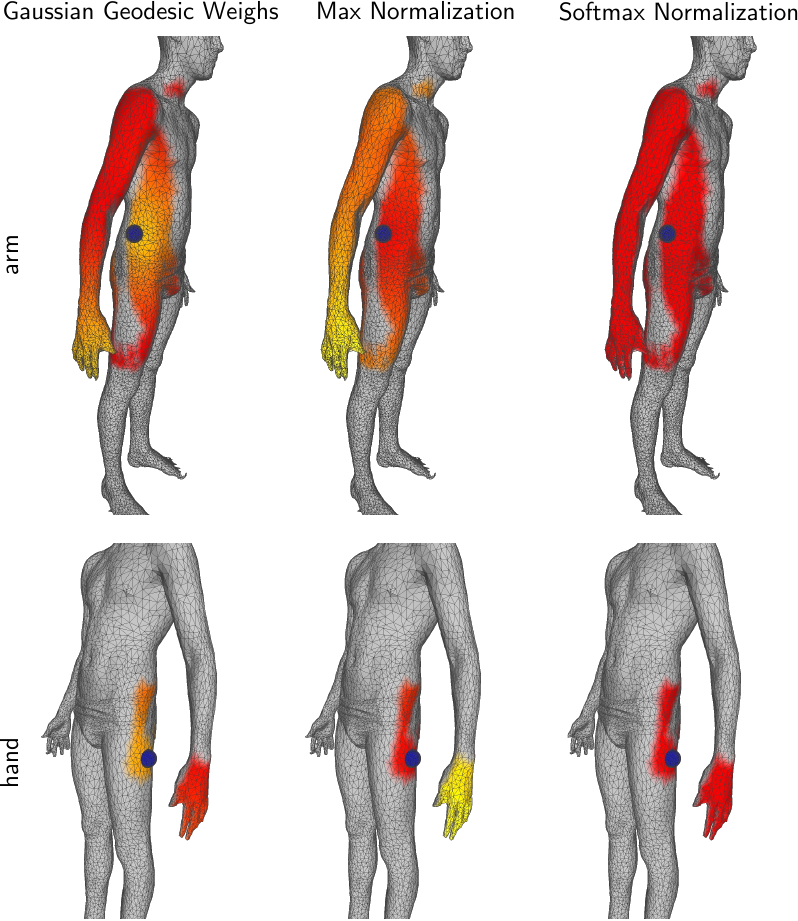}
    \caption{Gaussian Geodesic Weights can help reduce the scores for triangles inside a detected bounding box that do not belong to the segment of the corresponding text prompt.}
    \label{fig:normalizationheatmaps}
\end{figure}

%% file: sections/4_Experiments.tex
\section{Experiments}\label{sec:experiments}

\input{figures/faust_becnhmark}

\subsection{Experimental Setup}

\input{tables/faust_fg}

\input{tables/faust_cg}

\best{Datasets and splits.} 
We evaluate the zero-shot performance on our proposed FAUST \cite{FAUST} benchmark and on the ShapeNetPart ~\cite{Yi16} dataset. For the FAUST dataset, we use all of the 100 scans available. We manually collected the coarse and fine-grained annotations in the following way. First, we manually annotate the registered mesh of one human scan as shown in \Cref{fig:benchmark} using vertex paint in Blender \cite{Blender}. Since we have access to the FAUST mesh correspondences, we are able to transfer our annotations to all other scans. We then re-mesh each human scan independently to contain around 20K triangular faces. To generate annotations for these re-meshed scans, we assign the label of each vertex to the label of the nearest vertex before remeshing. For the ShapeNetPart dataset \cite{Yi16}, it contains 16 object categories and 50 annotated parts for shapes from the ShapeNet \cite{ShapeNet} dataset. We use all of the 2460 labeled shapes of the ShapeNetPart dataset provided by \cite{Kalogerakis20163DSS}, where the point labels are transferred to mesh polygon labels via a nearest neighbors approach combined with graph cuts. The meshes in ShapeNetPart have triangular faces that are very large, covering a large portion of the mesh surface. For this reason, during model inference, we provide re-meshed ShapeNetPart shapes as input, where each mesh contains at most 30K triangular faces. We use the Isotropic Explicit Remeshing algorithm \cite{Hoppe1993MeshO} provided by MeshLab \cite{LocalChapterEvents:ItalChap:ItalianChapConf2008:129-136} to do the re-meshing. During the evaluation, we transferred the predicted face labels back to the original ShapeNetPart shapes. \Cref{fig:teaser} includes examples from the TextANIMAR \cite{le2023textanimar}, Objaverse \cite{Deitke2022ObjaverseAU}, and TOSCA \cite{Bronstein2009NumericalGO} datasets.

\best{Metrics.} We use the semantic segmentation mIoU as described in \cite{PartNet}. We first calculate the mIoU for each part category across all the test shapes and then compute for each object category the average of the part mIoU.

\subsection{Implementation Details}
We use a single Nvidia V100 GPU for each experiment. We use the Nvidia Kaolin library \cite{KaolinLibrary} written in PyTorch \cite{NEURIPS2019_9015} for rendering in all of our experiments. To ensure fairness, we use the same views in all of our GLIP-based model experiments. 
As a pre-processing step, we center the input mesh around the origin and normalize it inside a unit sphere. For rendering, we use a resolution of $1024\times1024$ and a black background color.

\subsection{Zero-Shot Semantic Segmentation}

\subsubsection{FAUST Benchmark} 
We compare our method with 3DHighlighter \cite{3D_highlighter} and CLIPSeg \cite{CLIPSeg}. To obtain semantic segmentation results from 3DHighlighter, we run the optimization separately to get a highlighted mesh for each of the input semantic regions. If a face were highlighted for different semantic regions, its predicted label would be the semantic class with the highest CLIP similarity score.  To obtain semantic segmentation results from CLIPSeg, we generate a segmentation mask for each rendered view for each semantic label, and we aggregate the segmentation scores for each face and assign the most predicted label for each face.

In \Cref{tab:faustcg}, we report the overall average mIoU and the mIoU for each semantic part on our proposed FAUST benchmark. SATR significantly outperforms 3DHighlighter on the coarse-grained parts.  As shown in \Cref{fig:qualitative}, our method SATR outperforms 3DHighlighter and CLIPSeg on all of the four semantic parts (leg, arm, head, and torso) by an overall average mIoU of 82.46\%. In addition, we show that our proposed components(SATR) help improve the results upon our GLIP-based baseline (SATR-Baseline). In \Cref{fig:qualitative}, we show the qualitative results of our method, and we compare it to 3DHighlighter. In addition, in \Cref{tab:faustfg}, our method SATR overall outperforms 3DHighlighter with a margin of 42.12\% in the fine-grained semantic segmentation. In \Cref{fig:qualitative}, we show the results of SATR on fine-grained segmentation compared to the ground truth.

\input{figures/qualitative_results}
\input{tables/shapenetpart}

\subsubsection{ShapeNetPart Dataset}
In \Cref{tab:shapenetpart_results}, SATR consistently outperforms 3DHighlighter in every shape category by a large margin. The results suggest that our proposed method SATR is applicable not only to human shapes but can also perform well in a wide range of categories. In \Cref{fig:qualitative}, we compare SATR and 3DHighlighter. The main challenge is that 3DHighlighter only works with the right random initialization, which doesn't happen too often.

\subsection{Ablation Studies}

\noindent\best{Effectiveness of the proposed components.} We ablate the effectiveness of our proposed components for FAUST coarse and fine-grained benchmarks in \Cref{tab:ablationsFaustFG,tab:ablationsFaustCG}. Using both Gaussian geodesic re-weighting and visibility smoothing gave the best performance in both the coarse and fine-grained FAUST benchmarks. In addition, each component is effective on its own and performs better than SATR-Baseline. The results suggest that both components work in a complementary fashion.

\input{tables/sam_comparison}
\input{tables/components_ablations_coarse}
\input{tables/different_normalizations}

\noindent\best{Different re\-weighting methods.} We compare using different re-weighting methods as shown in \Cref{tab:ablationsnormalization}. As discussed earlier in \Cref{sec:method:geodesic}, we compute the geodesic distances between every visible face in a predicted bounding box and the capital face. To compute the weights for every visible face, we try re-weighting by doing the following $w_i=(1 - dist_i/(max_{dist} + \epsilon))$. We also try computing the weights by normalizing the distance with a softmax function. Our proposed Gaussian geodesic re-weighting method outperforms other methods, especially in the fine-grained benchmark with a very large margin, showing that it is robust when the capital face is miscalculated.

\input{figures/fine_grained_parts}

\noindent\best{Comparison using recent 2D segmentation models.} Recent foundation models for 2D semantic segmentation show promising results for zero-shot 2D semantic segmentation. For instance, SAM \cite{segmentAnything} can be combined with powerful 2D object detectors (like GLIP \cite{GLIP} and Grounding-DINO\cite{GroundingDINO}) for text-based semantic segmentation. In \Cref{tab:satrSAM}, we compare our proposed method with DINO-SAM and GLIP-SAM based segmentation methods. Our proposed object detector-based method still exhibits strong performance among recent works, especially in the FAUST fine-grained benchmark. 

\input{tables/components_ablation_fine}

%% file: figures/faust_becnhmark.tex
\begin{figure}[!htb]
    \centering
    \includegraphics[width=\linewidth]{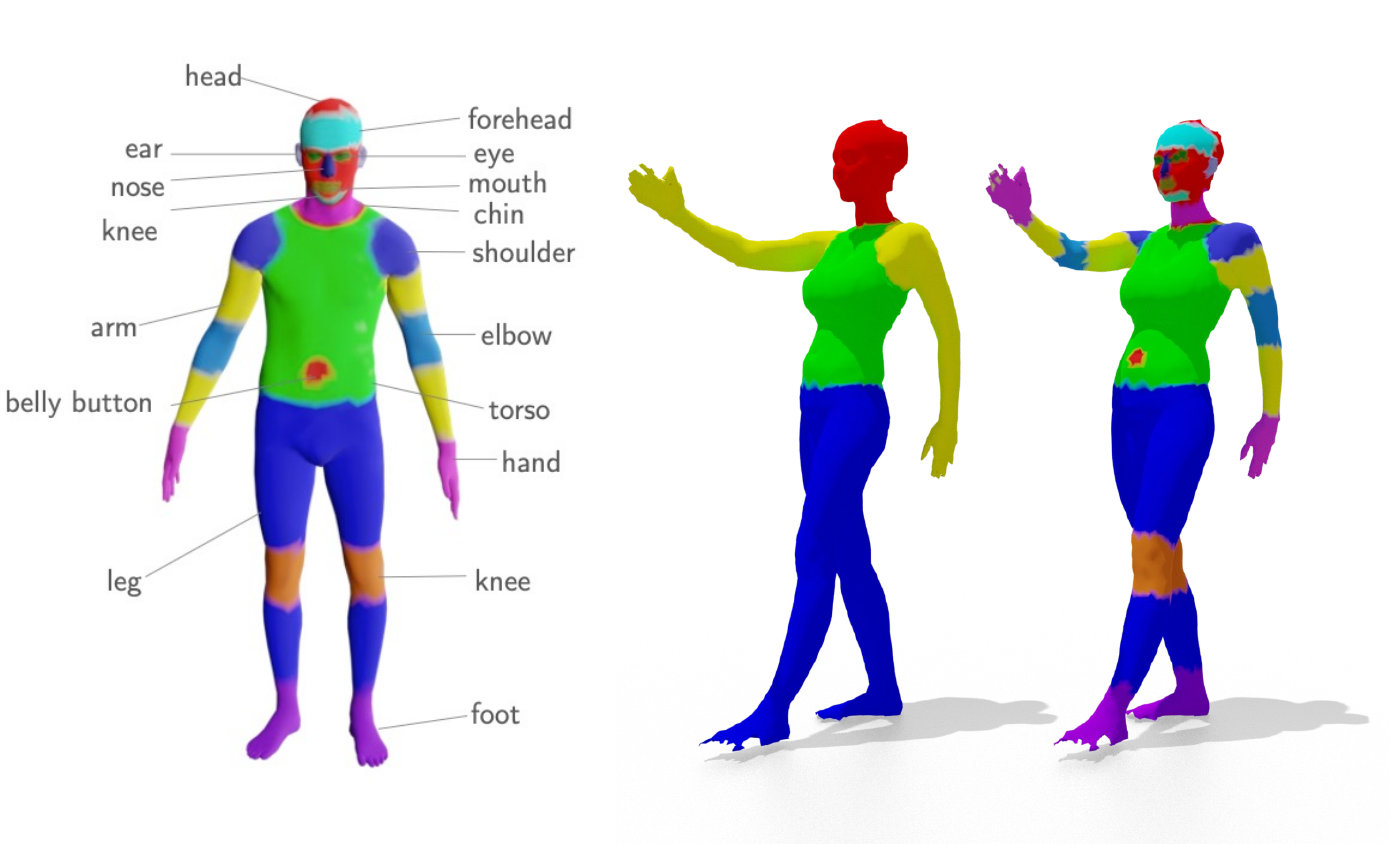}
    \caption{Proposed FAUST Benchmark. It contains 100 human shapes with both coarse and fine-grained annotations. 
    }
    \label{fig:benchmark}
\end{figure}

%% file: tables/faust_fg.tex
\begin{table*}[htb!]
\centering
\small
\setlength{\tabcolsep}{1.9pt}
\resizebox{1.0\linewidth}{!}{ 
\begin{tabular}{llccccccccccccccccccc}
\toprule
Method & Backbone & mIoU & arm &	\tabincell{c}{{belly}\\{button}}	& chin &	ear &	elbow &	eye &	foot &	\tabincell{c}{{fore}-\\{head}} &	hand &	head &	knee&	leg&	mouth&	neck&	nose&	\tabincell{c}{{shou-}\\{lder}}&	torso \\
\midrule

    3DH \cite{3dhighlighter} & CLIP \cite{CLIP} & 3.89 & 18.39& 1.99& 0.46& 0.72& 0.08& 0.0& 20.81& 0.70& 0.02& 3.49& 6.17& 3.91& 0.05& 1.94& 0.07& 0.04& 7.28 \\

\hline
    
    
\multirow{2}{*}{SATR-F}    & {CLIPSeg} &10.88 & 11.51 & 0.10 & 0.30 & 0.0 & 0.03 & 0.0 & 03.28 & 0.0 & 25.80 & 39.99 & 0.07 & 50.52 & 0.0 & 0.05 & 0.0 & 5.11 & 48.24 \\

    & {GLIP} & 41.96 & 45.22 &26.30 &\textbf{37.68} &41.67 &24.93 &\textbf{25.95} & 53.94 &\textbf{41.63} &68.22 &42.56 &32.69 &59.73 &\textbf{27.59} &41.78 &50.57 &33.00 &59.83 \\

\hline

    \multirow{1}{*}{SATR} & {GLIP} & \textbf{46.01} & \textbf{50.51} & \textbf{29.41} & {27.74} & \textbf{47.45} & \textbf{26.80} & 18.90 & \textbf{81.99} & 38.11 & \textbf{81.45} & \textbf{51.11} & \textbf{33.34} & \textbf{65.22} & 27.29 & \textbf{41.95} & \textbf{57.60} & \textbf{38.94} & \textbf{64.35} \\ 
\bottomrule
\end{tabular}
}
 \caption{Performance of SATR on the fine-grained semantic segmentation on FAUST dataset.}
 \label{tab:faustfg}
\end{table*}

%% file: tables/faust_cg.tex
\begin{table*}[!htb]
\centering

\begin{tabular}{llccccc} 
\toprule
 & Backbone & mIoU  & arm & head & leg & torso \\
\midrule
3DH \cite{3dhighlighter} &  \centering{CLIP \cite{CLIP}} & 16.50  & 28.60 &  14.20 &  14.90 &  8.20 \\
\hline


\multirow{3}{*}{SATR-F} & {LSeg \cite{LSeg}} & 6.50 & 26.00 & 0.0  & 0.0 & 0.0 \\

& {CLIPSeg \cite{CLIPSeg}} & 60.34 & 46.55 & 58.01 & 76.22 & 59.80 \\ 

& {GLIP \cite{GLIP}} & 81.16 & 82.01 & 88.17 & \textbf{86.54} & \textbf{67.92} \\ 

\hline
\multirow{1}{*}{SATR} & {GLIP \cite{GLIP}} & \textbf{82.46} & \textbf{85.92} & \textbf{90.56} & {85.75} & {67.60} \\

\bottomrule
\end{tabular}


\caption{Performance of SATR on the coarse-grained semantic segmentation on FAUST dataset.}
\label{tab:faustcg}

\end{table*}

%% file: figures/qualitative_results.tex
\begin{figure*}[!htb]
    \centering
    \includegraphics[width=\linewidth]{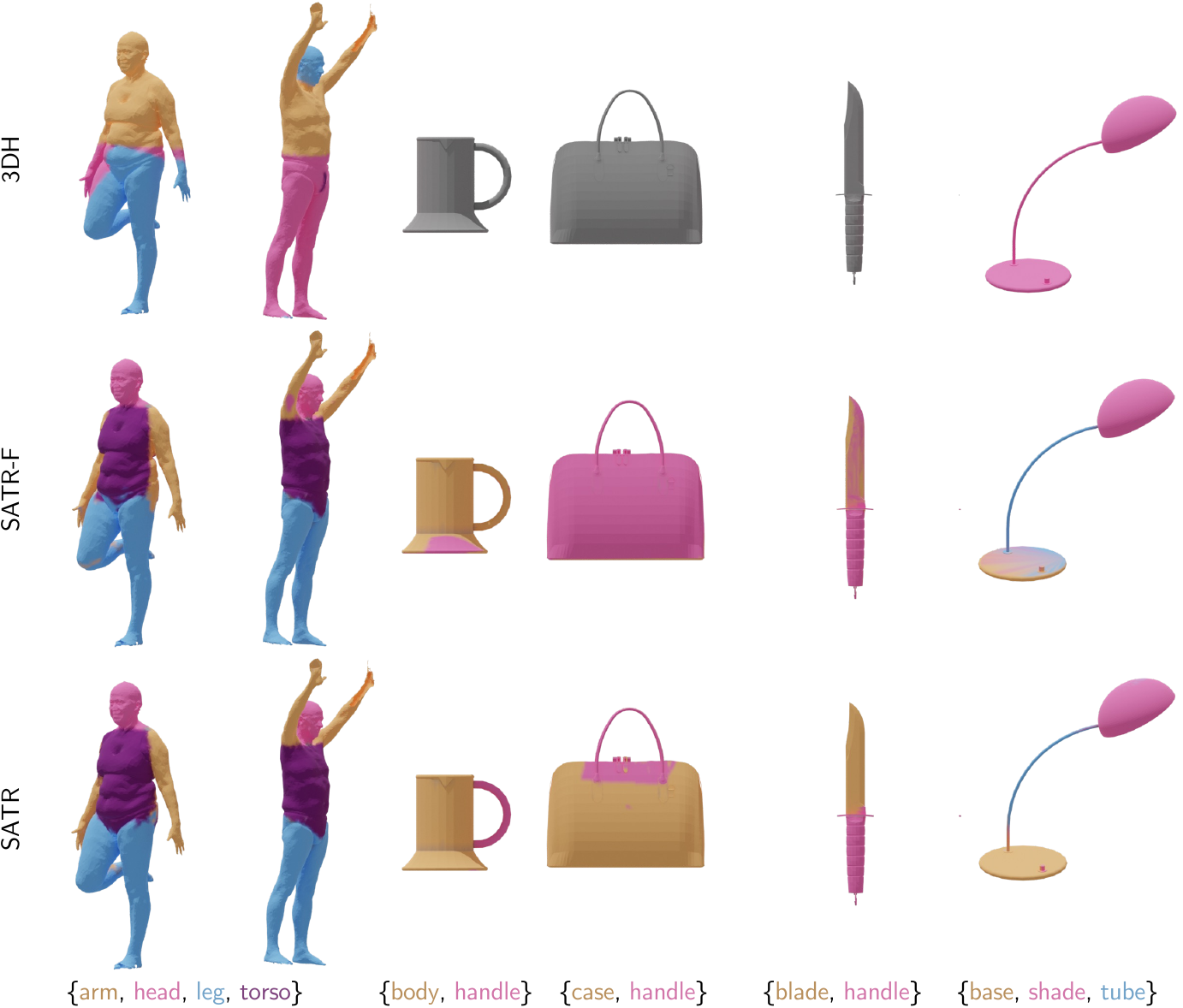}
    \caption{Qualitative results and comparison between 3DHighlighter, SATR-F, and SATR
    }
    \label{fig:qualitative}
\end{figure*}

%% file: tables/shapenetpart.tex
\begin{table*}[htb!]
\centering
\small
\setlength{\tabcolsep}{1.9pt}
\resizebox{1.0\linewidth}{!}{ 
\begin{tabular}{lcccccccccccccccccc}
    \toprule
    Method   & Backbone           & \tabincell{c}{mIoU}  & \tabincell{c}{air-\\plane} & bag & cap & car & chair & \tabincell{c}{ear-\\phone} & guitar & knife & lamp & laptop & \tabincell{c}{motor-\\bike} & mug & pistol & rocket & \tabincell{c}{skate-\\board} & table \\
    \midrule
    3DH \cite{3dhighlighter} & CLIP \cite{CLIP} & 5.70 & 5.81 & 2.05 & 2.85 & 2.88 & 15.53 & 9.55 & 0.86 & 1.58 & 13.21 & 1.78 & 5.57 & 0.65 & 1.36 & 10.36 & 6.44 & 10.77\\
    \hline
    {SATR-F} & GLIP \cite{GLIP} & 26.67 & \textbf{41.73} &26.60 &22.96 &\textbf{22.01} &{26.61} &14.95 &\textbf{43.55} &30.79 &\textbf{31.16} &30.05 &12.40 &31.55 &19.63 &15.55 &\textbf{34.49} &22.70 \\
    {SATR} & GLIP \cite{GLIP} &\textbf{31.90} & 38.46 &\textbf{44.56} &\textbf{24.01} &19.62 &\textbf{33.16} &\textbf{16.90} &40.22 &\textbf{45.92} &30.22 &\textbf{37.79} &\textbf{15.70} &\textbf{52.31} &\textbf{20.87} &\textbf{28.41} &30.77 &\textbf{31.41}  \\
    \bottomrule
\end{tabular}
}
\caption{Performance of SATR on the ShapeNetPart dataset.}
\label{tab:shapenetpart_results}
\end{table*}

%% file: tables/sam_comparison.tex
\begin{table}[!htb]
\centering
\resizebox{1.0\linewidth}{!}{ 
\begin{tabular}{lcccc} 
\toprule
  Method  & \tabincell{c}{Backbone(s)} & \tabincell{c}{Coarse-Grained\\ mIoU} &   \tabincell{c}{Fine-Grained\\ mIoU} \\
\midrule
\multirow{3}{*}{SATR-F} & GLIP \cite{GLIP}   & 81.16  & 41.96 \\
  & GLIP-SAM \cite{kirillov2023segment}  &  {78.59}  & {27.52} \\
  & DINO-SAM \cite{DINO}  &  {80.42}  & {21.90} \\
\hline

\multirow{1}{*}{SATR} & GLIP \cite{GLIP}&\textbf{82.46} & \textbf{46.01} \\

\bottomrule
\end{tabular}
}
\caption{mIoU comparison between SATR and SATR-F with recent SAM-based \cite{segmentAnything} backbones.\vspace{-3mm}}
    \label{tab:satrSAM}
\end{table}

%% file: tables/components_ablations_coarse.tex
\begin{table}[!htb]
\centering
\small
\resizebox{1.0\linewidth}{!}{ 
\begin{tabular}{ccccccc} 
\toprule
\tabincell{c}{{Gaussian Geod-}\\{esic Reweighting}}  & \tabincell{c}{{Visibility}\\{Smoothing}}  &  mIoU  & arm & head & leg & torso \\
\midrule
&  & 81.16 & 82.01 & 88.17 & 86.54 & 67.92 \\ 

& \checkmark & 81.69 & 82.68 & 88.61 & 86.85 & 68.61 \\

\checkmark & & 82.39 & 85.73 & \textbf{90.61} & \textbf{85.81} & \textbf{67.41} \\

\checkmark & \checkmark & \textbf{82.46} & \textbf{85.92} & {90.56} & {85.75} & {67.60} \\
\bottomrule
\end{tabular}
}
\caption{Ablation to show the effectiveness of our proposed components for the coarse-grained FAUST benchmark.}
\label{tab:ablationsFaustCG}
\end{table}

%% file: tables/different_normalizations.tex
\begin{table}[!htb]
\centering
\resizebox{1.0\linewidth}{!}{ 
\begin{tabular}{lcc} 
\toprule
Re-weighting Method  &  \tabincell{c}{{Coarse-Grained}\\{mIoU}} & \tabincell{c}{{Fine-Grained}\\{mIoU}} \\
\midrule
Max Geodesic & 82.41 & 44.57\\
Softmax Geodesic & 81.69  & 43.34\\
Gaussian Geodesic (ours) & \textbf{82.46} 
& \textbf{46.01}  \\
\bottomrule
\end{tabular}
}
\caption{Ablation on using different re-weighting methods. Our proposed Gaussian Geodesic Re-weighting method outperforms other normalization methods. This shows its effectiveness in the fine-grained and more difficult semantic segmentation task.}
\label{tab:ablationsnormalization}
\end{table}

%% file: figures/fine_grained_parts.tex
\begin{figure}[!htb]
    \centering
    \includegraphics[width=1.0\linewidth]{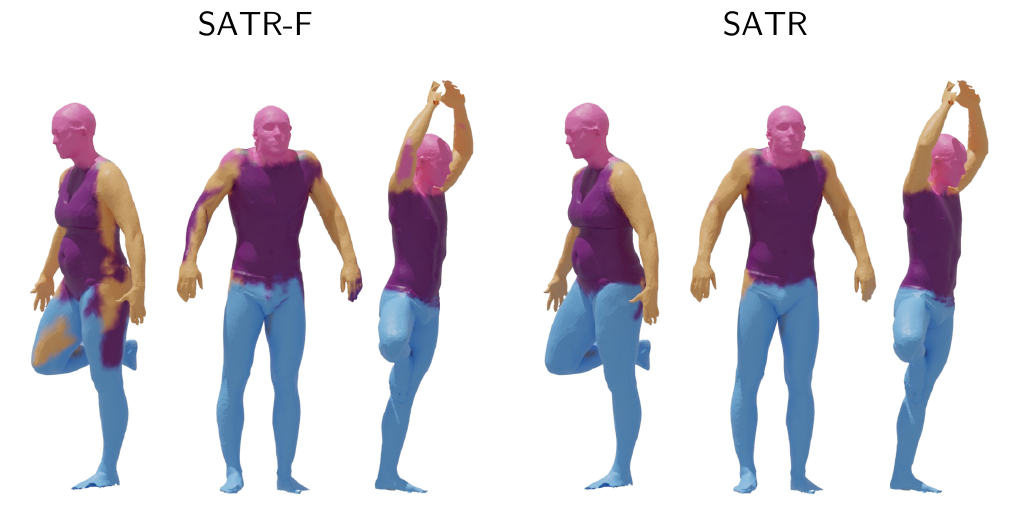}
    \caption{Segmentation leaking problem. Some semantically unrelated parts of the surface get incorrectly attributed to a given textual prompt because they often fall into predicted bounding boxes from multiple views.}
    \label{fig:leaking}
\end{figure}

%% file: tables/components_ablation_fine.tex
\begin{table}[htb!]
\centering
\small
\begin{tabular}{ccc}
\toprule
\tabincell{c}{{Gaussian Geodesic}\\{ Reweighting}}  & \tabincell{c}{{Visibility}\\{Smoothing}}    & mIoU  \\
\midrule
&   & 41.96 \\

& \checkmark  & {43.35}\\

\checkmark  &   & {45.56}  \\

\checkmark  & \checkmark  & \textbf{46.01} \\

\bottomrule
\end{tabular}
\caption{Ablation on trying all the possible combinations of our proposed components for the fine-grained FAUST benchmark.}
\label{tab:ablationsFaustFG}
\end{table}

%% file: sections/5_Conclusion.tex
\section{Conclusion}\label{sec:conclusion}

In our work, we explored the application of modern zero-shot 2D vision-language models for zero-shot semantic segmentation of 3D shapes.
We showed that modern 2D object detectors are better suited for this task than text-image similarity or segmentation models, and developed a topology-aware algorithm to extract 3D segmentation mapping from 2D multi-view bounding box predictions.
We proposed the first benchmarks for this area. We compared to previous and selected concurrent work qualitatively and quantitatively and observed a large improvement when using our method.
In future work, we would like to combine different types of language models and investigate in what way the names of the segments themselves can be proposed by a language model.
We discuss the limitations of our approach in Appx~\apref{sec:limitations}.

%% file: appendices/limitations.tex
\section{Limitations}\label{sec:limitations}

In this section, we discuss the limitations of our approach. First, there is no formal guarantee that the random view sampling algorithm covers each triangle. We mainly used this choice to be compatible with previous work. Alternatively, one could devise a greedy view selection algorithm that ensures each triangle is selected once. Another approach would be to use a set of uniformly distributed viewpoints that are fixed for the complete dataset. Second, we are not able to test other types of large language models because the latest versions are not publicly available \cite{GLIP2}. Third, GLIP prediction is not perfect. For example, it can incorrectly detect parts that are semantically different but still look similar (for instance, the front and the backpack of an astronaut, see \Cref{fig:glipFailureCase}).

%% file: appendices/implementation-details.tex
\section{SATR Pseudocode}\label{ap:pseudocode}
In Algorithm \ref{alg:satr}, we show the pseudocode of our proposed method (SATR). 

\input{algorithms/satr}

%% file: algorithms/satr.tex
\begin{algorithm}
\normalsize
\SetAlgoLined
\SetKwInOut{Input}{Input}\SetKwInOut{Output}{Output}
\Input{Zero-shot 2D object detector $D(\bm x, \bm t)$, where $x$ is an image and $t$ is a text prompt.}
\Input{Shape surface as a mesh of faces $\mathcal{F}$.}
\Input{Set of textual prompts $\mathcal{T}$ representing semantic regions/classes.}
\Input{Number of views $N_{views}$}
\Input{The q-ring neighborhood $q$.}
\Output{The predicted semantic label of each face of the input mesh $\mathcal{F}$.}

\highlight{\# Initialize face scores}  \\
$S$ = zeros($\mathcal{F}$.faces.length, $\mathcal{T}$.length), 

\highlight{\# Compute the pair-wise geodesic distance between every pair of faces}  \\
$G$ = \textit{computeFacePairwiseDistance}($\mathcal{F}$)

\highlight{\# Find the $q$-ring neighborhood for all the faces of the mesh} \\
$Q$ = \textit{getFaceQNeighbors}($\mathcal{F}$, q) \\

\highlight{\# Render the mesh} \\
$V$, \textit{Pixel2Face} = \textit{renderMesh}($\mathcal{F}$, $N_{views}$) \\

\For{$v$ in $V$}{
    \For{$t$ in $\mathcal{T}$} {
        \highlight{\# Detect 2D Bounding Boxes for the given $t$ prompt} \\
        $B_{v,t}$ = \textit{predictBoxes}($v$, $t$) \\
        \For{$b_{t, v}$ in $B_{v,t}$} {
            \highlight{\# Get the visible faces inside $b_{t, v}$} \\
            $f_{t, v}$ = \textit{getVisibleFaces}($b_{t, v}$, $\mathcal{F}$, \textit{Pixel2Face}) \\
            
            \highlight{\# Compute capital face} \\
            $c$ = \textit{computeCapitalFace}($f_{t, v}$, $\mathcal{F}$)

            \highlight{\# Compute frequency of the visible faces} \\
            $w_{freq}$ = faceFreq($f_{t, v}$, \textit{Pixel2Face})
            
            \highlight{\# Compute Gaussian Geodesic weights} \\
            $w_{geo}$ = \textit{faceGeodesicWeights}($c$, $f_{t, v}$)

            \highlight{\# Compute visibility smoothing weights} \\
            $w_{vis}$ = \textit{faceVisibilityWeights}($f_{t, v}$, $Q$)

            \highlight{\# Update face scores} \\
            $S_{f_{t, v}, i_t}$ +=  $w_{geo}$ * $w_{vis}$ * $w_{freq}$ \\
        }
    }
}
face\_label = \textit{argMax}($S$, axis=1) \\
Return face\_label\;
\caption{\methodfullname\ (\methodname) in high\-level pseudocode.}
\label{alg:satr}
\end{algorithm}

%% file: appendices/qualitative-results.tex
\begin{table}[!htb]
\centering
\resizebox{1.0\linewidth}{!}{ 
\begin{tabular}{lcc} 
\toprule
 &  \tabincell{c}{{Coarse-Grained}\\{mIoU}} &   \tabincell{c}{{Fine-Grained}\\{mIoU}} \\
\midrule
Uniform Sampling & 81.58 & 43.76 \\
Sampling using Normal Distribution \cite{3dhighlighter}  & 
\textbf{82.46} & \textbf{46.01} \\
\bottomrule
\end{tabular}
}
\caption{Ablation on using different view-port sampling methods in our proposed method SATR on FAUST coarse and fine-grained benchmarks. (with ten rendered views as input).}
\label{tab:ablationsCameraUniformSampling}
\end{table}

%% file: appendices/ablation-studies.tex
\section{Additional Ablation Studies}

\subsection{Changing the Number of Views}
We investigate the effect of changing the number of input rendered views on our proposed method SATR. In \Cref{tab:ablationsNViews}, we report the mIoU performance on both the coarse and the fine-grained FAUST benchmarks. Generally, increasing the number of views results in better segmentation performance in both benchmarks. However, the computation time increases, and there are diminishing returns for adding a large number of views.

\subsection{Camera View-port Sampling Approach}
We ablate using different sampling approaches for choosing the camera view-ports. In \Cref{tab:ablationsCameraUniformSampling}, we compare between sampling camera view-ports using the approach described in \cite{3D_highlighter} and doing uniform viewpoint sampling using a range of equidistant elevation and azimuth angles. We report the mIoU performance on both the coarse and the fine-grained FAUST benchmarks. We observe that sampling from a normal distribution ($\mu=3.14$, $\sigma=4$) results in better performance in both benchmarks. The reason behind this is having more control over the range of the elevation and azimuth angles can produce better views that cover most of the input mesh and avoid sampling views where a lot of occlusions may happen. More complex view selection could be future work.

\begin{figure}
    \centering
    \includegraphics[width=0.8\linewidth]{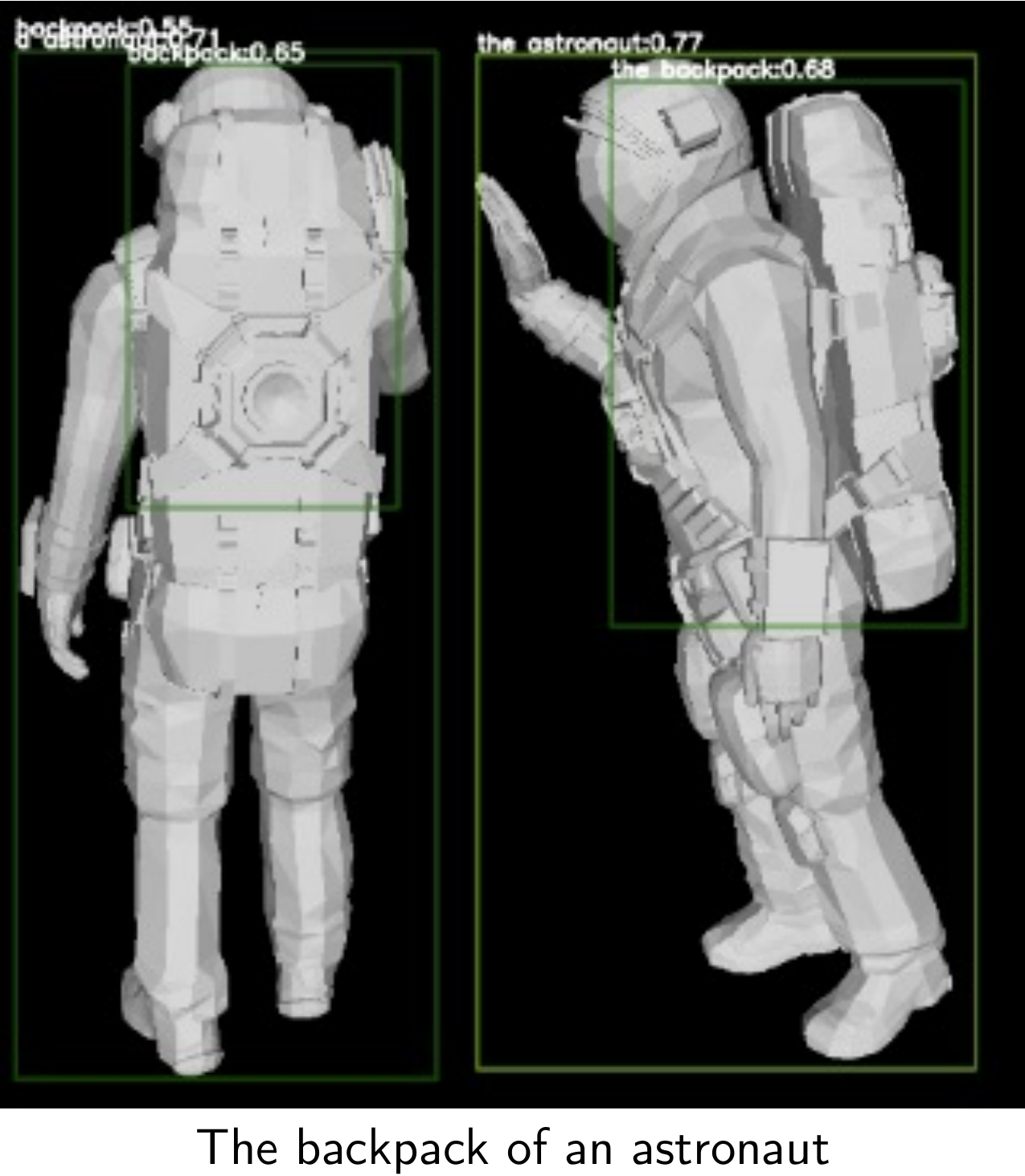}
    \caption{GLIP model can incorrectly detect semantically different parts that look similar. Since the parts of the astronaut suit look similar, GLIP mistakenly predicts the astronaut's chest as a part of the backpack and doesn't predict a tighter bounding box around the backpack.}
    \label{fig:glipFailureCase}
\end{figure}

\begin{table}[!htb]
\centering
\begin{tabular}{lcc} 
\toprule
 &  \tabincell{c}{{Coarse-Grained}\\{mIoU}} &   \tabincell{c}{{Fine-Grained}\\{mIoU}} \\
\midrule
Red & 80.62 & 40.67\\
Blue & 81.35 & 42.58\\
Skin color &82.22& 44.58\\
Gray & \textbf{82.46}& \textbf{46.01}\\
\bottomrule
\end{tabular}
\caption{Ablation on changing the vertex colors of the input 3D models to our proposed method SATR on FAUST coarse and fine-grained benchmarks. (with ten rendered views as input).}
\label{tab:ablationsCols}
\end{table}

\begin{table}[!htb]
\centering
\begin{tabular}{lcc} 
\toprule
\# Views&  \tabincell{c}{{Coarse-Grained}\\{mIoU}} &   \tabincell{c}{{Fine-Grained}\\{mIoU}} \\
\midrule
5 & 53.99 & 25.96 \\
10  & 82.46 & 24.3 \\
15  & 80.48 & 43.40 \\
20  & 82.41 & 45.20 \\
30  & 84.06 & \textbf{47.87} \\
40  & \textbf{84.26} & 47.67 \\

\bottomrule
\end{tabular}

\caption{Ablation on using a different number of rendered views as input to our proposed method SATR on FAUST coarse and fine-grained benchmarks.}

\label{tab:ablationsNViews}
\end{table}

\subsection{Coloring The Input Mesh}
We ablate the effect of changing the color of the input mesh on the performance of SATR. We run three experiments by using four different colors for the input meshes; grey, red, blue, and natural skin color. As shown in \Cref{tab:ablationsCols}, we find that using the gray color results in the best performance compared to other colors. 

\begin{table}[!htb]
    \centering
\resizebox{1.0\linewidth}{!}{ 
\begin{tabular}{llcc} 
\toprule
 & Backbone & Coarse-mIoU & Fine-mIoU \\
\midrule
SATR-F& {GLIP \cite{GLIP}} & 81.16 & 41.96 \\ 
\multirow{1}{*}{SATR (add)} & {GLIP \cite{GLIP}} & \textbf{82.75} & {44.90} \\
\multirow{1}{*}{SATR (mul)} & {GLIP \cite{GLIP}} & 82.46 & \textbf{46.01} \\

\bottomrule
\end{tabular}
}    
    \caption{Ablation on different ways of combining the reweighting factors.  }
    \label{tab:addition_of_factors_ablation}
\end{table}

\subsection{Summation of Reweighting Factors}
We replace the multiplication of both of the Gaussian Geodesic and Visiblity Smoothing reweighting factors as shown in \Cref{eq:face_score_computation} with addition:

\begin{equation}
\weights_m^*[n, k] = \sum_{\ell=1}^{L_{m, k}}{\weights^{\ell}_m[n, k, \ell] \times \area_{m,\ell}^n \times \bm (r_\ell^{m,k} + \bm \vis^{m,k}[n]}). 
\end{equation}

We show the mIoU performance on both FAUST coarse and fine-grained benchmarks in \Cref{tab:addition_of_factors_ablation}. The addition of the reweighting factors gave slight increase in performance for the coarse benchmark while perform significantly worse than multiplying the 
reweighting factors.